\pdfoutput=1
\documentclass[sigconf]{acmart}
\usepackage{stfloats}
\newcommand{\tabincell}[2]{\begin{tabular}{@{}#1@{}}#2\end{tabular}}
\usepackage{tabularx}
\usepackage{microtype}
\usepackage{hhline}  
\usepackage{multirow}

  \usepackage{amsmath,amssymb,amsfonts}
\usepackage{algorithmic}
\usepackage{graphicx}
\usepackage{textcomp}
\usepackage{xcolor}
\usepackage{multirow}
\usepackage{makecell}
\usepackage{subfigure}
\usepackage{epstopdf}
\usepackage{caption}

\usepackage{amsthm,amsmath,amssymb}
\usepackage{mathrsfs}

\AtBeginDocument{%
  \providecommand\BibTeX{{%
    \normalfont B\kern-0.5em{\scshape i\kern-0.25em b}\kern-0.8em\TeX}}}

\setcopyright{acmcopyright}
\copyrightyear{2021}
\acmYear{2021}
\acmDOI{10.1145/1122445.1122456}

\acmConference[MM '21]{}{October 20--24, 2021}{Chengdu, China}
\acmBooktitle{MM '21, October 20--24, 2021, Chengdu, China}
\acmPrice{15.00}
\acmISBN{978-1-4503-XXXX-X/18/06}

\begin{document}

\title{GeneAnnotator: A Semi-automatic Annotation Tool\\ for Visual Scene Graph}



\author{Zhixuan Zhang}
\affiliation{%
  \institution{School of Software Engineering, Xi’an Jiaotong University}
  \city{Xi'an}
  \country{China}
  }

\author{Chi Zhang}
\affiliation{%
  \institution{College of Artificial Intelligence, Xi'an Jiaotong University}
  \city{Xi'an}
  \country{China}
  }
  
\author{Zhenning Niu}
\affiliation{%
  \institution{College of Artificial Intelligence, Xi'an Jiaotong University}
  \city{Xi'an}
  \country{China}
  }
  
\author{Le Wang}
\affiliation{%
  \institution{College of Artificial Intelligence, Xi'an Jiaotong University}
  \city{Xi'an}
  \country{China}
  }

\author{Yuehu Liu}
\authornotemark[1]
\affiliation{%
  \institution{College of Artificial Intelligence, Xi'an Jiaotong University}
  \city{Xi'an}
  \country{China}
  }
\email{liuyh@mail.xjtu.edu.cn}

\begin{abstract}
In this manuscript, we introduce a semi-automatic scene graph annotation tool for images, the GeneAnnotator. This software allows human annotators to describe the existing relationships between participators in the visual scene in the form of directed graphs, hence enabling the learning and reasoning on visual relationships, e.g., image captioning, VQA and scene graph generation, etc. The annotations for certain image datasets could either be merged in a single VG150 data-format file to support most existing models for scene graph learning or transformed into a separated annotation file for each single image to build customized datasets. Moreover, GeneAnnotator provides a rule-based relationship recommending algorithm to reduce the heavy annotation workload. With GeneAnnotator, we propose Traffic Genome, a comprehensive scene graph dataset with 1000 diverse traffic images, which in return validates the effectiveness of the proposed software for scene graph annotation. The project source code, with usage examples and sample data is available at https://github.com/Milomilo0320/A-Semi-automatic-Annotation-Software-for-Scene-Graph, under the Apache open-source license.
\end{abstract}




\captionsetup{aboveskip=0.1\normalbaselineskip}
\begin{teaserfigure}
  \centering
  \includegraphics[width=\textwidth]{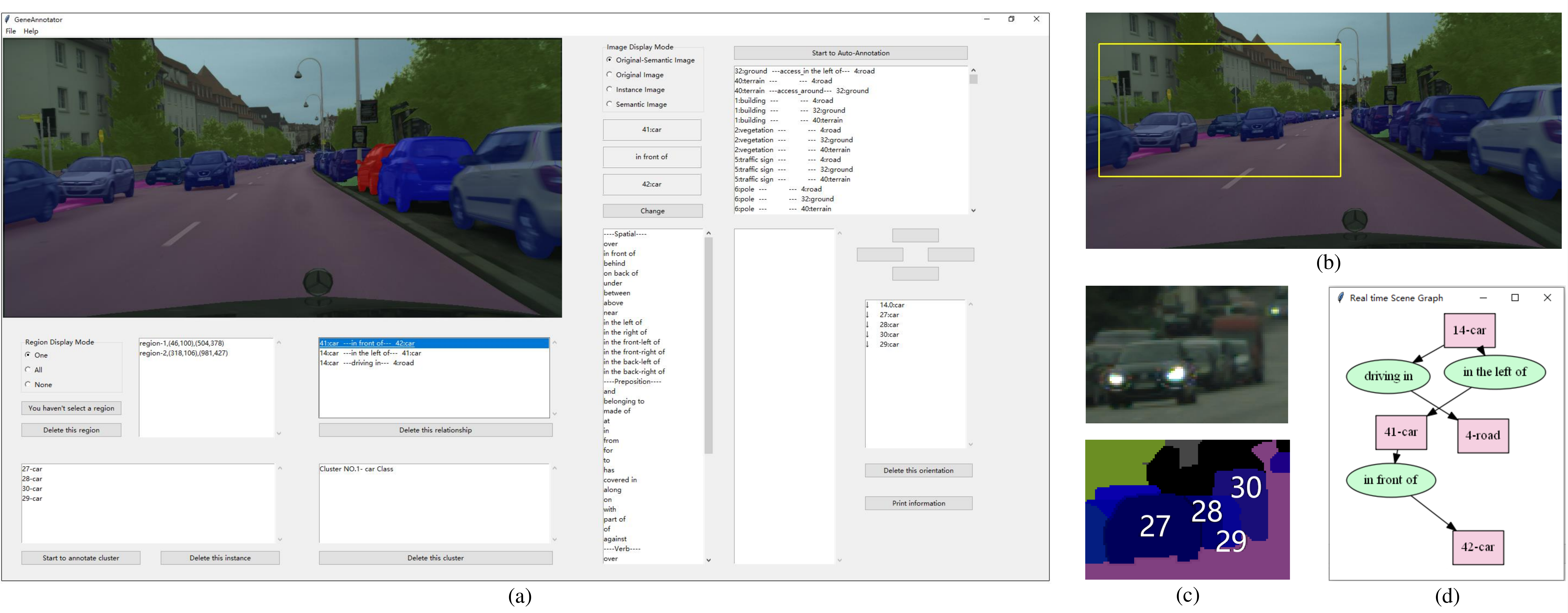}
  \caption{Screenshots of GeneAnnotator. (a) The user interface. The relationship between the subject, i.e. the red-masked car, and the object, i.e. the blue-masked car in the image region is defined by a directed graph, and described by editing or selecting from the recommended candidate relationships listed on the GUI. (b) To avoid trivial and irrelevant relationships, a “region” of interest, i.e. the yellow bounding box, could be set during the annotation. (c) Nearby cars can be annotated as a “cluster” in GeneAnnotator, sharing the same relationship with other cars or roads. (d) The real-time visualized scene graph according to annotations in (a).\\}
  \label{fig:teaser}
\end{teaserfigure}

\maketitle

\begin{figure*}[h]
  \centering
  \includegraphics[width=18cm]{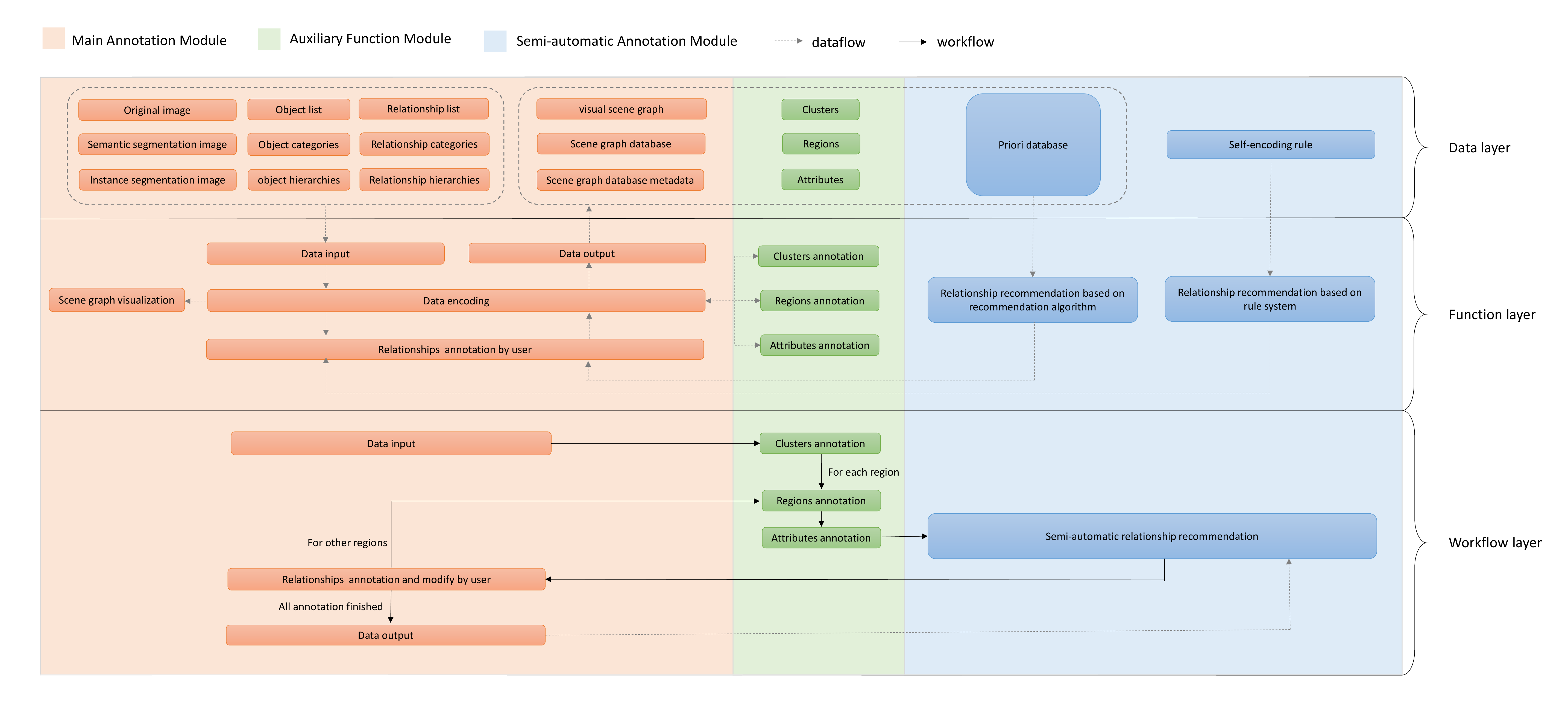}
  \caption{System Architecture. In GeneAnnotator, there are three core modules: Main Annotation Module, Semi-automatic Annotation Module and Auxiliary Function Module. Each module is independent of one another. Removing one of the modules does not affect the use of the whole software, which makes it easy for developers to update and maintain.}
  \label{fig:SA}
\end{figure*}

\section{Introduction}
Scene graph is a topological structured data representation of visual content. It is usually represented by a directed graph, the nodes of which represent the instances and the edges represent the relationship between instances. Specifically, for a relationship, the starting node is called the subject, and the ending node is called the object. In recent years, scene graph has been applied to semantic image retrieval, image captioning, visual question answering and achieved reasonable performance.

Large-scale scene graph datasets with precise manually annotated are the prerequisites for related research. The open-source annotation tool is relatively uncommon at present. Existing datasets cannot organize data into a unified data framework. It is not conducive to customize datasets.

In order to fill these gaps, we propose the GeneAnnotator, a semi-automatic annotation software for scene graph generation by Python. It has the following major features: \textbf{(1) Friendly user interactivity.} Figure \ref{fig:teaser} shows the GUI of GeneAnnotator. Users can visually view the contents annotated and scene graph in real time. \textbf{(2) Diverse information.} In addition to the relationships between instances, it also allows users to annotate diverse information such as regions, clusters and attributes. \textbf{(3) Semi-automatic annotation.} GeneAnnotator can recommend the relationship according to the recommendation algorithm and rule system, which greatly improves the annotation efficiency. \textbf{(4) Highly customized.} Users are free to define the underlying data, including object categories, relationship lists and so on. It is helpful for researchers to build customize datasets. \textbf{(5) Applying easily.} The output dataset files are organized in VG150's data format and can be ready to most existing models for scene graph learning.

At the same time, we provide Traffic Genome, a scene graph dataset with 1000 traffic images built by GeneAnnotator. Traffic Genome contains information about the location, labels, relationships, attributes and other information of the elements in traffic scenes. It is one of the 
earliest open-source traffic scene graph datasets. In comparison to other related datasets, such as Visual Genome\cite{krishna2017visual} and VG150\cite{2017Scene}, objects and relationships in Traffic Genome are denser and the attribute coverage is highest.

\section{System Architecture}

In this section, we present the system architecture of GeneAnnotator, as illustrated in Figure \ref{fig:SA}. We first detail the functionalities and concepts in core modules and then introduce the annotation workflow.

\subsection{Main Annotation Module}

Main Annotation Module is the most basic module. Its main functions include data input and output, data encoding, relationship annotation and scene graph visualization. This module allows users to freely define the underlying data, i.e. object lists, object categories, object hierarchies, relationship lists, relationship categories, and relationship hierarchies. 

\textbf{Object hierarchy and relationship hierarchy\cite{10.1145/3436369.3437437}:} objects and relationships are hierarchical and instances connect to others within or across different layers to form a scene graph.

\subsection{Auxiliary Function Module}

Auxiliary Function Module allows researchers to annotate regions, clusters and attributes.

\textbf{Region:} In order to make the scene graph directly express the visual perception of images, we suggest user located each visually meaningful region with a bounding box like Figure \ref{fig:teaser}(b). The region is helpful to avoid annotating trivial and irrelevant relationships in scenes.

\textbf{Cluster:} Users can regard objects nearby and attributes as a "cluster". For example, nearby cars in Figure \ref{fig:teaser}(c) were annotated as a “cluster”. The practice has proved that clusters can make the annotation more efficient.

\textbf{Attributes:} Attributes are an additional description of objects that can provide finer-grained nodes for the scene graph to better distinguish different instances. In GeneAnnotator, “orientation” attributes are constructed for the research content. The researchers can add the required attributes to software according to their own needs, such as speed, color, size and so on.

\begin{table*}
  \caption{A comparison among Traffic Genome, Visual Genome and VG150.}
  \label{tab:comparison}
  \resizebox{!}{1.1cm}{
  \begin{tabularx}{17.6cm}{ccccccc}
    \toprule
    Datasets & \tabincell{c}{\quad Images} & \tabincell{c}{Object\\\quad Categories} & \tabincell{c}{Total\\ \quad \quad Instances} & \tabincell{c}{Total Instances\\ \quad in Scene Graph}& \tabincell{c}{Percentage of Instances\\in Scene Graph} & \tabincell{c}{Instances in Scene\\Graph per Image}  \\
    \midrule
    \textbf{Traffic Genome}   & 1,000   & 34     & 25,146     & 19,291    & \textbf{76.71\%}  & \textbf{19.29} \\
    Visual Genome & 108,077 & 33,877 & 3,843,636  & 2,254,357 &  58.65\% & 20.85\\
    VG150         & 108,072 & 150    & 1,145,397  & 614,625   &  53.66\% & 5.69\\
    \bottomrule
  \end{tabularx}}
  
  \resizebox{!}{1.1cm}{
  \begin{tabularx}{17.6cm}{cccccccc}
    \toprule
    Datasets & \tabincell{c}{Relationship\\Categories} & \tabincell{c}{Total\\Relationships} & \tabincell{c}{\textls[-35]{Relationships}\\\textls[-35]{per Image}} & \tabincell{c}{\textls[-35]{Relationships per} \\\textls[-35]{Instance in Scene Graph}}& \tabincell{c}{Attribute\\Categories} & \tabincell{c}{Instances \\\textls[-35]{with Attribute}} & \tabincell{c}{Attribute\\Coverage}  \\
    \midrule
    \textbf{Traffic Genome}    & 51    & 29,191    & \textbf{29.19}  & \textbf{3.02}    & 4       & 25146    & \textbf{100\%} \\
    Visual Genome & 36,550 & 1,531,448 & 14.17  & 1.36   
    & 52,577  & 1670182  & 42.72\%\\
    VG150         & 50     & 622,704  & 5.76    & 2.02    & 200     & 720496   & 62.90\%\\
    \bottomrule
  \end{tabularx}}
\end{table*}

\subsection{Semi-automatic Annotation Module}

Semi-automatic Annotation Module recommends the best relationship for each pair of instances according to the recommendation algorithm and rule system.

\textbf{Recommendation algorithm:} Based on Simple Tag-based Recommendation algorithm, <subject, relationship, object> is represented as $(u, i, b)$. $u$ indicates a pair of <subject, object>, which has relationship $i$ and attribute $b$. $b$ is a vector, each element in which is a visual attribute of the pair of $u$, like “whether the profile touch” or “whether the different size of bounding boxes”. The equation of interest between a pair of <subject, object> $u$ and their relationship $i$ is given by:

\begin{equation}
  P(u,i)=\sum{n_{u,b}n_{b,i}},
\label{eq:1}
\end{equation}

where $n_{u,b}$ indicates the number of times that $u$ has attribute $b$. $n_{b,i}$ indicates the number of times that $u$ with attribute $b$ has relationships $i$.

The prior database consists of the annotated relationships. The parameters in Equation (\ref{eq:1}) can be quickly updated and relationships are recommended without training.

\textbf{Rule system:} The recommendation algorithm is based on a large number of prior information. In the stage of start in the sparse data, self-encoding rule are used to guess a relationship. The default rules in GeneAnnotator is to judge the relationship according to the contact and position of two instances.

\subsection{Workflow}

The workflow of building a scene graph dataset is shown in the workflow layer in Figure \ref{fig:SA}. First of all, the configuration files of the original image and other underlying data are read into GeneAnnotator. GUI shows the image and user on-demand region annotation and cluster annotation. Then the Semi-automatic Annotation Module recommends relationships, which users can modify or add to. The annotated relationships will be stored in the prior database to enrich the recommendation algorithm. Finally, the output files are the scene graph dataset.

\section{Dataset}

\subsection{Specifications}
Traffic Genome is a traffic scene graph dataset comprised of 1000 traffic scenes. The original images are selected from Cityscapes\cite{cordts2016cityscapes}, which is a pixel-level semantic segmented dataset.

We offer two different formats of Traffic Genome. One format is following the structure of VG150, which covers all the scene graphs (1000 images). According to related work, most of the existing scene graph learning models use the database and metadata proposed in VG150. Therefore, files in this format can be directly used in existing models. However, it does not support users to modify or add relationships to a specified scene graph. Thus, we designed another format for per-image, containing image instances map, clusters annotations and other more detailed information. It is convenient to modify the specified scene graph. We also provide code to convert between the two formats.

\subsection{Statistics}
In Traffic Genome, there are 34 semantic object classes (including "unlabeled") and 51 relationships. In comparison to related datasets, objects and relationships in Traffic Genome are denser and the attribute coverage is highest. As shown in Table \ref{tab:comparison}, there is a comparison among Traffic Genome, Visual Genome and VG150.

\textbf{Object density:} In Traffic Genome, there are 25,147 annotated instances, of which 19,291 are involved in the scene graph. On average per image, Traffic Genome has 19.29 instances were involved in a scene graph, which is very close to the 20.85 in Visual Genome and much larger than the 5.69 in VG150. At the same time, the percentage of instances involved in a scene graph in Traffic Genome is about 76.71\%, much larger than the 58.65\% of Visual Genome and the 53.66\% of VG150. It indicates that Traffic Genome has richer and finer annotations, allowing more instances in the images to participate in the construction of the scene graph.

\textbf{Relationship density:} We annotated 29,192 relationships in Traffic Genome. On average, each scene graph has 29.19 relationships, much larger than the 14.17 in Visual Genome and the 5.76 in VG150. We also calculated the average of relationships involved per object. The results show that one object has relationships with another 3.02 objects in Traffic Genome. The value in Visual Genome and VG150 is 1.36 and 2.02. It can be regarded as a measure of the sparsity of a scene graph. A denser graph has more edges involved per node. That means Traffic Genome has a much denser relationship annotation than the other two datasets.

Most of the relationships are focused on spatial relationships (such as “in left of”) and area relationships(such as “driving on”), 43.85\% and 42.04\% of the total relationships, respectively. The top 5 common combination of <Object – Relationship - Subject> are : “Person – Walking on - Sidewalk”, “Car – In front of/In back of - Car”, “Car – driving on - Road”, “Car – Parking on - Road”, “Person – Walking on - Road”.

\textbf{Attribute coverage:} Traffic Genome annotated oriented attribute for each instance, i.e. "forward", "leftward", "rightward" and "backward". The attribute coverage is 100\%, which is the highest of the three datasets.

\subsection{Applications and Evaluations}
We use Traffic Genome to experiment and evaluate the following scene graph generation models:

\begin{itemize}

\item IMP\cite{2017Scene}: The model solves the scene graph inference problem using standard RNNs and learns to iteratively improves its predictions via message passing.
\item Motif\cite{2017Neural}: Regularly appearing substructures in scene graphs make object labels are highly predictive of relation labels.
\item VC-Tree\cite{2020Learning}: Proposing a method to compose dynamic tree structures that place the objects in an image into a visual context.
\item VCTree-TDE\cite{2020Unbiased}: Building a causal graph and making biased training with the graph.
\item EBM-Loss\cite{suhail2021energybased}: The additional constraint in the learning framework acts as an inductive bias and allows models to learn efficiently from a small number of labels.
\end{itemize}

Since Traffic Genome was given the labels of ground-truth, we only evaluated the performance of predicate classification. Models successfully run with Traffic Genome. As shown in Table \ref{tab:result}, the predicate classification results of Traffic Genome are very close to VG150. It even performs better in mR@100. By evaluating Traffic Genome in scene graph generation models, GeneAnnotator is proved to be a valuable tool for scene graph annotation.

\begin{table}[htbp]
\caption{The predicate classification results on different models.}
\begin{center}
\resizebox{!}{3cm}{
\begin{tabular}{c|c|c|c|c|c}
    \toprule
    \multirow{2}*{Model} & \multirow{2}*{Method} & \multirow{2}*{Dataset} &\multicolumn{3}{c}{Predicate Classification} \\
    \cline{4-6}
    & & & \textls[-35]{mR@20} & \textls[-35]{mR@50} & \textls[-55]{mR@100} \\
    \hhline{======} 
    
    \multirow{4}*{IMP}
    & \multirow{2}*{\textls[-55]{Cross Entropy}}& VG150 & 8.85 & 10.97 & 11.77\\
                        &  & \textls[-35]{Traffic Genome} & 7.03 & 11.62 &15.44 \\
    \cline{2-6}
    & \multirow{2}*{\textls[-35]{EBM-Loss}}& VG150 & 9.43 & 11.83 & 12.77\\
                        &  & \textls[-35]{Traffic Genome} & 3.50 & 5.98 &7.89 \\
    \hhline{======}                     
    \multirow{4}*{Motif}
    & \multirow{2}*{\textls[-55]{Cross Entropy}}& VG150 & 12.45 & 15.71 & 16.8\\
                        &  & \textls[-35]{Traffic Genome} & 12.13 & 18.55 &21.64 \\
    \cline{2-6}
    & \multirow{2}*{\textls[-35]{EBM-Loss}}& VG150 & 14.17 & 18.02 & 19.53\\
                        &  & \textls[-35]{Traffic Genome} & 11.73 & 17.33 &19.80 \\
    \hhline{======}                     
    \multirow{4}*{VCTree}
    & \multirow{2}*{\textls[-55]{Cross Entropy}}& VG150 & 13.07 & 16.53 & 17.77\\
                        &  & \textls[-35]{Traffic Genome} & 13.48 & 19.79 &24.85 \\
    \cline{2-6}
    & \multirow{2}*{\textls[-35]{EBM-Loss}}& VG150 & 14.20 & 18.19 & 19.72\\
                        &  & \textls[-35]{Traffic Genome} & 11.37 & 17.92 &20.87 \\
    \hhline{======} 
    \multirow{4}*{\tabincell{c}{VCTree\\-TDE}}
    & \multirow{2}*{\textls[-55]{Cross Entropy}}& VG150 & 16.30 & 22.85 & 26.26\\
                        &  & \textls[-35]{Traffic Genome} & 12.82 & 18.33 &22.79 \\
    \cline{2-6}
    & \multirow{2}*{\textls[-35]{EBM-Loss}}& VG150 & 19.87 & 26.66 & 29.97\\
                        &  & \textls[-35]{Traffic Genome} & 10.30 & 15.32 &19.71 \\
    
    \bottomrule
\end{tabular}}
\label{tab:result}
\end{center}
\end{table}

\section{Avaliability}
GeneAnnotator is released under the license of Apache 2.0 and its source code is openly available at: https://github.com/Milomilo0320/A-Semi-automatic-Annotation-Software-for-Scene-Graph. In addition, Traffic Genome will be made available at https://github.com/Milomi\\lo0320/Traffic-Scene-Graph-1000. Contributions from the open-source community are welcome, via the GitHub issues/pull request mechanisms.

\section{Conclusion}
In this paper, we have developed GeneAnnotator, a semi-automatic annotation software for scene graph generation by Python. It is one of the earliest open-source scene graph annotation tools. With modular designs, GeneAnnotator is easy-to-use and easy-to-extend. Furthermore, we provide Traffic Genome, a scene graph dataset with 1000 traffic images built by GeneAnnotator. By evaluating Traffic Genome in scene graph generation models, GeneAnnotator is proved to be a valuable tool for scene graph annotation. 

\begin{acks}
This work was supported by the National Key Research and Development Program of China, No. 2018AAA0102504.
\end{acks}

\bibliographystyle{ACM-Reference-Format}
\bibliography{sample-base}

\end{document}